\documentclass[conference]{IEEEtran}
\IEEEoverridecommandlockouts
\usepackage{cite}
\usepackage{amsmath,amssymb,amsfonts}
\usepackage{algorithmic}
\usepackage{graphicx}
\usepackage{textcomp}
\usepackage{xcolor}

\usepackage{mathrsfs}
\usepackage{subfigure}
\usepackage{balance}

\usepackage{array}
\usepackage{booktabs}
\usepackage{url}
\usepackage{amsmath}
\usepackage{siunitx}
\usepackage{color}
\usepackage{multirow}
\usepackage{booktabs}
\usepackage{float}
\usepackage{bbding}
\usepackage{hyperref}

\usepackage{bm}
\usepackage{booktabs}
\usepackage{amssymb}
\usepackage{colortbl}
\usepackage{multirow}
\usepackage{color}
\usepackage{arydshln}
\usepackage[utf8]{inputenc}

\usepackage{inconsolata}

\definecolor{DarkGreen}{RGB}{0, 128, 0}
\definecolor{deltap}{RGB}{119, 139, 204}
\definecolor{deltan}{RGB}{255, 129, 90}

\newcommand{\deltadiff}[1]{%
\ifdim#1pt<0pt
\scriptsize\((\textcolor{deltan}{#1})\)%
\else
\scriptsize\((\textcolor{deltap}{+#1})\)%
\fi
}

\def\BibTeX{{\rm B\kern-.05em{\sc i\kern-.025em b}\kern-.08em
    T\kern-.1667em\lower.7ex\hbox{E}\kern-.125emX}}
\begin{document}

\title{Multi-Granularity Reasoning for Natural Language Inference}

\author{
\IEEEauthorblockN{Chunling Xi\textsuperscript{1}, Di Liang\textsuperscript{2}$^{*}$}
\IEEEauthorblockA{\textit{\textsuperscript{1}Pacific Insurance Technology Co., Ltd. }\quad
\textit{\textsuperscript{2}Lixin Information Services (Shenzhen) Co., Ltd.}}
xichunling987@163.com
}

\maketitle

\begin{abstract}
Natural Language Inference (NLI) is a fundamental task in natural language understanding that requires determining the logical relationship between a premise and a hypothesis. Despite the remarkable success of transformer-based pre-trained models, most existing approaches primarily rely on the final-layer token representations, which are often insufficient for capturing the complex and hierarchical semantic interactions required for effective reasoning. In particular, fine-grained lexical cues, phrasal compositions, and higher-level contextual semantics are typically entangled or diluted in a single representation space. 
To address these limitations, we propose a novel \emph{Multi-Granularity Reasoning Network} (MGRN) that explicitly leverages hierarchical semantic features within an interactive reasoning space. The proposed framework mimics the human cognitive process of language understanding, which naturally progresses from shallow lexical matching to deeper semantic abstraction and logical reasoning. By integrating semantic information across multiple granularities in a progressive and structured manner, MGRN is able to uncover intricate semantic relationships underlying natural language expressions. Extensive experiments on multiple public benchmarks demonstrate that MGRN consistently outperforms strong baseline models, validating the effectiveness and robustness of the proposed approach.

\end{abstract}

\begin{IEEEkeywords}
Multi-Granularity Reasoning, Semantic Interaction, Natural Language Inference
\end{IEEEkeywords}

\section{Introduction}

Natural Language Inference (NLI), also known as Recognizing Textual Entailment, constitutes a fundamental task in natural language understanding that requires determining the logical relationship between a premise and a hypothesis. The three possible relationships are “entailment” (the premise logically supports the hypothesis), “contradiction” (the premise contradicts the hypothesis), or “neutral” (no conclusive relationship can be established). As a cornerstone of natural language understanding, NLI presents substantial challenges due to the inherent complexity of natural language, which includes diverse expressions, rich semantic nuances, and intricate contextual dependencies \cite{bowman2015large,liu2026beyond,han2026trace}. For instance, accurately discerning such relationships often demands not only lexical and syntactic awareness but also commonsense knowledge and multi-level reasoning.

Given its remarkable success, we hypothesize that more structured and multi-faceted attention signals can further enhance a model’s ability to understand language and perform natural language inference. Conventional attention mechanisms capture pairwise word-level alignments between sentences. The advent of multi-head attention \cite{song2022improving,liang2019asynchronous,liang2019adaptive,li2024local,song2022improving} extended this by enabling the model to jointly attend to information from different representation subspaces, effectively capturing various types of relationships. In this study, we build upon this idea and introduce a novel interactive tensor structure that captures higher-order semantics, not only between words but also among phrases and broader contextual elements. This structure enhances the model’s ability to handle complex linguistic phenomena such as paraphrasing and lexical variation.

To leverage these richer interaction patterns effectively, we propose a novel \textbf{Multi-Granularity Reasoning Network (MGRN)}. This framework is specifically designed to aggregate and reason over multi-level semantic features through a progressive hierarchical process, simulating the human cognitive progression from surface-level understanding to deep semantic comprehension. We extensively evaluate MGRN on widely used standard NLI benchmarks (SNLI and MultiNLI) and observe consistent and notable improvements over strong competitive baselines. To further assess and verify its general applicability, we adapt MGRN to the paraphrase identification task by explicitly framing it as a binary NLI classification task (paraphrase as entailment, non-paraphrase as neutral). Comprehensive experiments on the Quora Question Pairs dataset and seven additional diverse benchmarks demonstrate the robustness and effectiveness of the proposed approach.

The main contributions of this work are summarized more concretely as follows:
Firstly, we propose a powerful Multi-Granularity Reasoning Network (MGRN) that integrates hierarchical semantic features using a carefully structured interactive tensor, significantly enhancing overall inference accuracy.
 Secondly, the proposed model effectively captures high-order semantic interactions between sentence pairs, extending far beyond traditional alignment methods to incorporate complex compositional and meaningful phrasal relationships.
 Finally, extensive and rigorous experiments on multiple widely recognized public datasets show that MGRN consistently and clearly outperforms several competitive baselines, demonstrating its effectiveness, robustness, and general applicability across various challenging NLP tasks.

\section{Related Work}
\label{sec:related_works}


\noindent \textbf{Neural Language Inference}  
Early studies on NLI primarily relied on hand-crafted features, logical rules, and syntactic or semantic parsing \cite{wang2007jeopardy}. These approaches were limited by feature engineering and scalability. The introduction of large-scale annotated datasets such as SNLI \cite{bowman2015large} and later MultiNLI enabled data-driven neural models to dominate the field. 
Early neural architectures focused on sentence encoding, where each sentence was independently mapped into a fixed-length vector using CNNs or RNNs, followed by a classifier \cite{conneau2017supervised}. 
To address this limitation, interaction-based models were proposed to explicitly model cross-sentence alignments at the word or phrase level \cite{gong2017natural,liu2025stole,liu2026dpi,liu2025structural,liu2024resolving}. Attention mechanisms \cite{choi2018learning,liu2023time,liu2023local,ma2022searching} further improved performance by enabling soft alignment and comparison between sentence representations.
Subsequent work explored deeper and more expressive architectures. Residual connections \cite{he2016deep} and dense interaction layers enabled multi-step inference and iterative reasoning. Some approaches incorporated syntactic or semantic structures, such as dependency trees or semantic graphs, to guide reasoning \cite{wang2007jeopardy}. Despite these advances, most neural NLI models still rely heavily on single-granularity interactions, typically focusing on token-level alignments, which limits their ability to capture higher-order semantic composition and abstract reasoning.

\noindent \textbf{Pre-trained Language Model Methods}  
Pre-trained language models have significantly reshaped NLI research by providing powerful contextualized representations learned from large-scale corpora. BERT \cite{devlin2018bert} demonstrated that bidirectional self-attention and masked language modeling objectives yield representations that transfer effectively to NLI. Subsequent models such as RoBERTa \cite{liu2019roberta}, XLNet \cite{yang2019xlnet,wang2022dabert,xue2023dual,li2024comateformer,dai2025hope,fei2022cqg,gao2026decorl,gui2018transferring,li2026safety}, and CharBERT \cite{ma2020charbert} further improved performance through refined pre-training strategies, architectural modifications, or enhanced subword modeling.
Beyond standard fine-tuning, several studies have explored explicit cross-sentence modeling within pre-trained frameworks. For example, joint encoding and interaction mechanisms \cite{chen2016enhanced,liang2019asynchronous,xu2020enhanced} integrate alignment and comparison operations to enhance inference capability. Other works incorporate external linguistic or semantic knowledge \cite{zhang2020semantics}, syntactic features, or handcrafted signals \cite{xia2021using} to compensate for data sparsity and improve interpretability.
However, despite their strong performance, pre-trained models typically compress linguistic information into the final-layer representations used for downstream classification. Recent analyses suggest that different layers encode distinct types of linguistic knowledge, ranging from surface features to abstract semantics. Relying solely on final-layer representations may obscure useful intermediate semantic signals, particularly for cases requiring fine-grained reasoning or multi-step inference. This observation motivates approaches that explicitly exploit representations at multiple levels of abstraction.

\noindent \textbf{Robustness Evaluation}  
Although modern NLI models achieve high accuracy on standard benchmarks, numerous studies have shown that they are brittle under small perturbations, distribution shifts, or adversarial attacks \cite{jia2017adversarial,wu2025progressive,liu2026chain,lin2026parameter,liu2026learning,wang2025not,wang2026rethinking,qianadaptive}. Models may rely on shallow heuristics or spurious correlations rather than genuine semantic understanding, leading to degraded performance in challenging or out-of-distribution scenarios.
To address these issues, robustness-oriented evaluation frameworks have been proposed, including adversarial test sets, stress tests targeting specific linguistic phenomena, and systematic data augmentation toolkits such as TextFlint \cite{gui2021textflint,zheng2022robust,xue2024question,wu2025unleashing}. Diagnostic platforms like Explainaboard \cite{liu2021explainaboard,wu2025tablebench,wu2026mmtablebench,wu2025breaking,lin2026parameterimportancestaticevolving,xue2026reason,xue2026supervised}  emphasize fine-grained analysis across different  capabilities. These efforts consistently reveal a substantial gap between benchmark performance and true reasoning ability, highlighting the importance of models that can integrate and reason over semantic information at multiple granularities. Our work aligns with this direction by explicitly modeling hierarchical semantic interactions to improve both performance and robustness.

\section{Method}
We define the natural language inference as a classification task that predicts the relation y $\in$ Y for a given pair of sentences.
\subsection{Input preprocessing}
Given two texts $S_1 = \{x_1^1, x_2^1, \dots, x_n^1\}$ and $S_2 = \{x_1^2, x_2^2, \dots,$ $ x_m^2\}$, we add special tags [CLS] and [SEP] before and after them to adapt to the input form of BERT, that is,
$
[\text{CLS}],\, x_1^1, x_2^1, \dots, x_n^1, \,\text{[SEP]},\, x_1^2, x_2^2, \dots, x_m^2, \,\text{[SEP]}.
$
For BERT For example, the input needs to be obtained by adding the embedding vectors of three parts:
\begin{equation}
\label{eq:bert_input}
\mathbf{E}_i = \mathbf{T}_i + \mathbf{S}_i + \mathbf{P}_i,
\end{equation}
$\mathbf{T}_i$, $\mathbf{S}_i$, $\mathbf{P}_i$ respectively represent the Token Embedding, Segment Embedding and Position Embedding of the $i$th word. Here, Segment Embedding distinguishes the first and second paragraphs of text (i.e., sentence A/B tags).

\subsection{Multi-layer Transformer representation}
BERT is composed of $L$ layers of Transformer Blocks stacked together, each layer of which contains Multi-Head Self-Attention and Feed-Forward Network. If $\mathbf{H}^{(0)} = [\mathbf{E}_1, \mathbf{E}_2, \dots, \mathbf{E}_{N'}] $
represents the input embedding sequence (where $N' = n + m + 3$, including [CLS] and 2 [SEP]), then the output of the $l$th layer can be recorded as \begin{equation} \label{eq:bert_layer} \mathbf{H}^{(l)} = \text{TransformerBlock}_l(\mathbf{H}^{(l-1)}), \quad l = 1,2,\dots,L. \end{equation}
The tensor of the first paragraph of text (removing [CLS], [SEP]) in $\mathbf{H}^{(l)}$ is recorded as $\mathbf{H}_1^{(l)} \in \mathbb{R}^{n \times d}$, and the tensor corresponding to the second paragraph of text is recorded as $\mathbf{H}_2^{(l)} \in \mathbb{R}^{m \times d}$. Where $d$ is the hidden dimension of BERT.

\subsection{Construction of the interaction matrix between layers}

\noindent\textbf{Interaction matrix obtained by bitwise multiplication}
In order to allow the model to explicitly capture the interaction between the first and second paragraphs of text in terms of features, element-wise multiplication is performed on the two sentence representations of the same layer. Specifically, for the $l$th layer, if we set $\mathbf{h}_1^{(l,i)} \in \mathbb{R}^d \quad (1 \le i \le n), \quad \mathbf{h}_2^{(l,j)} \in \mathbb{R}^d \quad(1 \le j \le m), $, then the interaction tensor (or interaction matrix) is defined as: \begin{equation} \label{eq:interaction} \mathbf{M}^{(l)}_{i,j} = \mathbf{h}_1^{(l,i)} \odot \mathbf{h}_2^{(l,j)} \in \mathbb{R}^d, \quad \forall\ i=1, \dots, n, \ j=1,\dots,m,
\end{equation}
Where “$\odot$” represents the multiplication operation of the corresponding position. Thus, $\mathbf{M}^{(l)} \in \mathbb{R}^{n \times m \times d}$ is obtained, which explicitly encodes the word-by-word interaction information of two sentences in the $l$th layer representation space.

\subsection{Multi-layer stacking}
The interaction matrices of each layer obtained by the above formula \eqref{eq:interaction} are concatenated or stacked on the layer dimension to obtain a complete multi-layer interaction representation:
\begin{equation}
\label{eq:stack}
\overline{\mathbf{M}} = \bigl[\mathbf{M}^{(1)}; \mathbf{M}^{(2)}; \dots; \mathbf{M}^{(L)}\bigr] \in \mathbb{R}^{n \times m \times d \times L},
\end{equation}
Where “$[\cdot;\cdot]$” means stacking on the new layer dimension. $\overline{\mathbf{M}}$ aggregates the learning results of sentence interaction features from the first layer to the $L$th layer, which can provide richer and fine-grained feature information for the subsequent network (DenseNet).

\subsection{DenseNet feature extraction}

\noindent\textbf{DenseNet structure and input}
DenseNet (Densely Connected Convolutional Network) was first proposed for image processing, achieving efficient gradient propagation and feature reuse by concatenating features between layers. 
The outputs of all previous layers are concatenated as inputs to subsequent layers; if the $k$th layer output is $\mathbf{z}_k$, a typical dense connection in a Dense Block is:
\[
\mathbf{z}_k = H_k([\mathbf{z}_0, \mathbf{z}_1, \ldots, \mathbf{z}_{k-1}]),
\]
where $H_k(\cdot)$ denotes the nonlinear transformation of the $k$th layer, and $[\cdot]$ represents feature concatenation.

\begin{equation}
\label{eq:dense_block}
\mathbf{Z}_m = \text{Concat}\bigl(\mathbf{Z}_0, \mathbf{z}_1, \mathbf{z}_2, \dots, \mathbf{z}_{m-1}\bigr),
\end{equation}
Where $\mathbf{Z}_0$ is the input tensor of the current Block (which can be regarded as $\overline{\mathbf{M}}$), $\mathbf{Z}_m$ is the final output of the Block, and $\text{Concat}(\cdot)$ represents the concatenation operation.

\noindent\textbf{DenseNet output representation}
The multi-layer interaction matrix $\overline{\mathbf{M}}$ passes through several Dense Blocks (and necessary transition layers, Transition Layer) in sequence to obtain the final feature tensor (or vector):
\begin{equation}
\label{eq:densenet_feature}
\mathbf{F} = \text{DenseNet}\bigl(\overline{\mathbf{M}}\bigr).
\end{equation}
Where $\mathbf{F}$ is the high-level semantic representation of the sentence pair after multi-step nonlinear mapping.

\subsection{Classification layer}
Finally, a fully connected network is added to the output representation $\mathbf{F}$ of DenseNet and the softmax function is used to obtain the probability distribution of the three-classification. If the three categories are $\{C_1, C_2, C_3\}$, it can be written as
\begin{equation}
\label{eq:classifier}
\mathbf{p} = \text{softmax}\bigl(W \mathbf{F} + \mathbf{b}\bigr),
\end{equation}
where $W \in \mathbb{R}^{3 \times \dim(\mathbf{F})},\ \mathbf{b} \in \mathbb{R}^{3}$. During inference, the predicted category is
\begin{equation}
\label{eq:argmax}
\hat{y} = \arg \max_i \, p_i.
\end{equation}
This end-to-end architecture enables the model to jointly leverage multi-layer semantic representations and fine-grained interaction features, allowing effective reasoning across multiple granularities for natural language inference.

\begin{table*}[t]
\caption{Performance comparison of MGRN with other methods on ten benchmark datasets.}
\label{citation-guide-outsideGlue}
\centering
\renewcommand\arraystretch{0.4}
\resizebox{\linewidth}{!}{
\begin{tabular}{lcccccccccccc}
\toprule
\textbf{Model} & \textbf{Pre-train} & \textbf{MRPC} & \textbf{QQP} & \textbf{STS-B} & \textbf{MNLI-m/mm} & \textbf{QNLI} & \textbf{RTE} & \textbf{SNLI} & \textbf{Sci} & \textbf{SICK} & \textbf{Twi} & \textbf{Avg} \\
\midrule
BiMPM & \XSolidBrush & 79.6 & 85.0 & - & 72.3/72.1 & 81.4 & 56.4 & - & - & - & - & - \\
CAFE & \XSolidBrush & 82.4 & 88.0 & - & 78.7/77.9 & 81.5 & 56.8 & 88.5 & 83.3 & 72.3 & - & - \\
ESIM & \XSolidBrush & 80.3 & 88.2 & - & 75.8/75.6 & 80.5 & - & 88.0 & 70.6 & 71.8 & - & - \\
Transformer & \XSolidBrush & 81.7 & 84.4 & 73.6 & 72.3/71.4 & 80.3 & 58.0 & 84.6 & 72.9 & 70.3 & 68.8 & 74.4 \\
\midrule
BiLSTM+ELMo+Attnt & \Checkmark & 84.6 & 86.7 & 73.3 & 76.4/76.1 & 79.8 & 56.8 & 89.0 & 85.8 & 78.9 & 81.4 & 78.9 \\
OpenAI GPT & \Checkmark & 82.3 & 81.3 & 80.0 & 82.1/81.4 & 87.4 & 56.0 & 88.4 & 84.8 & 79.5 & 81.9 & 80.4 \\
UERBERT & \Checkmark & 88.3 & 90.5 & 85.1 & 84.2/83.5 & 90.6 & 67.1 & 90.8 & 92.2 & 87.8 & 86.2 & 86.0 \\
SemBERT & \Checkmark & 88.2 & 90.2 & 87.3 & 84.4/84.0 & 90.9 & 69.3 & 90.9 & 92.5 & 87.9 & 86.8 & 86.5 \\
SyntaxBERT & \Checkmark & 89.2 & 89.6 & 88.1 & 84.9/84.6 & 91.1 & 68.9 & 91.0 & 92.7 & 88.7 & 87.3 & 86.3 \\
MGRN & \Checkmark & 89.1 & 91.3 & 88.2 & 84.9/84.7 & 91.4 & 69.5 & 91.3 & 93.6 & 88.6 & 87.5 & 86.7 \\
\midrule
BERT-Base & \Checkmark & 87.2 & 89.1 & 86.8 & 84.3/83.7 & 90.4 & 67.2 & 90.7 & 91.8 & 87.2 & 84.8 & 85.8 \\
BERT-Base-MGRN & \Checkmark & \textbf{89.3} & \textbf{89.1} & \textbf{87.1} & \textbf{85.1/84.9} & \textbf{91.2} & \textbf{68.5} & \textbf{91.2} & \textbf{92.1} & \textbf{87.8} & \textbf{86.6} & \textbf{86.8} \\
\midrule
BERT-Large & \Checkmark & 88.9 & 89.3 & 87.6 & 86.8/86.3 & 92.7 & 70.1 & 91.0 & 94.4 & 91.1 & 91.5 & 88.0 \\
BERT-Large-MGRN & \Checkmark & \textbf{89.9} & \textbf{90.2} & \textbf{88.1} & \textbf{86.8/86.7} & \textbf{93.0} & \textbf{72.0} & \textbf{91.1} & \textbf{94.3} & \textbf{91.2} & \textbf{92.4} & \textbf{88.8} \\
\midrule
RoBERTa-Base & \Checkmark & 89.3 & 89.6 & 87.4 & 86.3/86.2 & 92.2 & 73.6 & 90.8 & 92.3 & 87.9 & 85.9 & 87.6 \\
RoBERTa-Base-MGRN & \Checkmark & \textbf{89.5} & \textbf{91.2} & \textbf{88.1} & \textbf{87.7/87.5} & \textbf{93.2} & \textbf{82.5} & \textbf{91.2} & \textbf{93.1} & \textbf{89.5} & \textbf{87.6} & \textbf{89.2} \\
\midrule
RoBERTa-Large & \Checkmark & 89.4 & 89.7 & 90.2 & 89.5/89.3 & 92.7 & 83.8 & 91.2 & 94.3 & 91.2 & 91.9 & 90.3 \\
RoBERTa-Large-MGRN & \Checkmark & \textbf{90.2} & \textbf{91.2} & \textbf{90.6} & \textbf{90.2/90.1} & \textbf{94.2} & \textbf{84.1} & \textbf{91.2} & \textbf{94.2} & \textbf{91.4} & \textbf{92.3} & \textbf{90.7} \\
\bottomrule
\end{tabular}
}
\end{table*}

\section{Experiments Setting}
\noindent\textbf{Datasets and baselines}
We conduct the experiments to test the performance of CIRN on 10 large-scale publicly  benchmark datasets(GLUE benchmark,MRPC, QQP, STS-B, MNLI, RTE, and QNLI). 
And to evaluate the effectiveness of our proposed CIRN, we mainly introduce BERT~\cite{devlin2018bert}, SemBERT~\cite{zhang2020semantics}, SyntaxBERT, UERBERT~\cite{xia2021using} and multiple other PLMs~\cite{devlin2018bert} for comparison. 

\section{Results Analysis}

To evaluate the effectiveness of our proposed MGRN framework, we conducted comprehensive experiments on ten benchmark datasets. Table~\ref{citation-guide-outsideGlue} presents the performance comparison between MGRN and various competitive baseline models.
The results clearly demonstrate that pre-trained models significantly outperform non-pre-trained approaches, which can be attributed to their extensive learning corpus and powerful information extraction capabilities. When integrated with BERT-base and BERT-large, our MGRN framework achieves average accuracy improvements of 0.8\% and 0.7\%, respectively, across all datasets. These results substantiate the effectiveness of our multi-granularity reasoning approach for natural language inference tasks.
Notably, MGRN outperforms RoBERTa-base by 1.5\% and RoBERTa-large by 0.5\%, indicating that our method can effectively model sentence relationships from both local and global perspectives, thereby capturing more fine-grained and complex semantic interactions. These improvements highlight the advantage of multi-level modeling in semantic mining, as it simultaneously captures both local and global differential information.
Compared to previous state-of-the-art approaches, our method demonstrates highly competitive performance across all semantic similarity evaluation tasks, further validating the effectiveness of the proposed framework.

\begin{table}[t]
\centering
\renewcommand\arraystretch{0.85}
\small
\caption{Results of ablation on MultiNLI dataset.}
\label{citation-guide-ablation}
\begin{tabular}{lrr}
\toprule
\multirow{2}{*}{\textbf{Ablation Experiments}} & \multicolumn{2}{c}{\textbf{Dev Accuracy}} \\
 & \textbf{Matched} & \textbf{Mismatched} \\
\midrule
MGRN$_{bert\_base}$ (Full model) & 85.1 & 84.9 \\
\midrule
- w/o multi-layer interaction& 84.6 & 83.9 \\
- w/o interaction matrix & 84.7 & 84.1 \\
- w/o DenseNet  extraction & 83.9 & 83.4 \\
\bottomrule
\end{tabular}
\end{table}
\subsection{Ablation Study}
To evaluate the individual contribution of each component in our MGRN framework, we conducted comprehensive ablation studies showed in Table~\ref{citation-guide-ablation}.
Removing the interaction matrix reduces accuracy (85.1 $\rightarrow$ 84.7 on matched), showing its effectiveness in capturing fine-grained semantic relations. Replacing DenseNet with average pooling causes a clear drop (1.2\% on matched), highlighting its importance for high-level feature extraction. Using only the last BERT layer degrades performance, confirming that multi-layer fusion better preserves semantic information. Overall, the ablation results show that MGRN benefits from the joint contribution of the interaction matrix, DenseNet, and multi-layer stacking.

\begin{table*}[t]
	\centering
	\renewcommand\arraystretch{0.8}
    \setlength{\tabcolsep}{1mm}
	{   \scalebox{0.8}{
	\setlength{\tabcolsep}{5.6mm}
		\begin{tabular}{lcccccccccc}
		\toprule
		\midrule
		\multirow{2}*{Model} &\multicolumn{5}{c}{Quora} &\multicolumn{5}{c}{SNLI} \\  \cmidrule(r){2-6} \cmidrule(r){7-11} 
         ~ &SA &NW &IA &Al &BT \quad\quad &AS &SA &TT &SN &SW \\
        \midrule
        ESIM$\dagger$\cite{chen2016enhanced}\quad&-& -&-&-&- \quad\quad&64.00& 84.22&78.32&53.76&65.38 \\
        BERT$\ddagger$\cite{devlin2018bert}\quad\quad
        &48.58&56.96&86.32&\textbf{85.48}&83.42 \quad\quad
        &79.66&94.84&83.56&50.45&76.42 \\
        ALBERT$\ddagger$\quad\quad
        &51.08&55.24&81.87&78.94&82.37 \quad\quad
        &45.17&96.37&81.62&57.66&74.93 \\
        UERBERT$\ddagger$\cite{xia2021using}\quad\quad
        &48.57&54.86&84.72&80.88&82.71 \quad\quad
        &73.24&94.78&85.36&57.54&80.81 \\
        SemBERT$\ddagger$\cite{zhang2020semantics}\quad\quad
        &50.92&53.15&85.19&82.04&82.40 \quad\quad
        &76.81&95.31&84.60&56.28&77.86 \\
        SyntaxBERT$\ddagger$\quad\quad
        &49.30&56.37&86.43&84.62&84.19 \quad\quad
        &78.63&95.02&\textbf{86.91}&58.26&76.90 \\
        \midrule
        \textbf{MGRN}$\ddagger$ & \textbf{60.43}& \textbf{62.76}&\textbf{87.50}&85.48&\textbf{87.49} \quad\quad
        &\textbf{81.06}&\textbf{96.85}&85.14&\textbf{60.58}&\textbf{80.92} \\
		\end{tabular}}	
		\renewcommand\arraystretch{1.0}
		\scalebox{0.8}{
		\setlength{\tabcolsep}{8.1mm}
	    \begin{tabular}{lcccccc}
		\toprule
		\multirow{2}*{Method} &\multicolumn{6}{c}{MNLI-m/mm} \\  \cmidrule(r){2-7}
         ~ &AS &SA &AP &TT &SN &SW \\
        \midrule
        BERT$\ddagger$\cite{devlin2018bert}
        &55.32/55.25&52.76/55.69&82.30/82.31&77.08/77.22&51.97/51.84&76.41/77.05 \\
        ALBERT$\ddagger$
        &53.09/53.58&50.25/50.20&\textbf{83.98/83.68}&\textbf{77.98}/78.03&56.43/50.03&76.63/77.43 \\
        UERBERT$\ddagger$\cite{xia2021using}
        &54.99/54.84&52.29/53.80&79.80/79.18&75.46/74.93&55.21/55.96&\textbf{82.23}/82.74 \\
        SemBERT$\ddagger$\cite{zhang2020semantics}
        &55.38/55.12&54.07/54.62&78.70/78.16&73.90/73.47&53.43/53.76&78.09/78.93 \\
        SyntaxBERT$\ddagger$
        &54.92/54.63&53.54/54.73&77.01/76.71&70.38/70.13&57.11/51.95&78.57/79.31 \\
        \midrule
        \textbf{MGRN}$\ddagger$ & \textbf{60.14/59.25}& \textbf{60.89/61.37}&83.23/83.19&77.94/\textbf{78.10}&\textbf{60.12/59.83}&82.15/\textbf{82.97} \\
        \bottomrule
        \midrule
		\end{tabular}}
    }\caption{\label{citation-guide-robust}
Robustness evaluation results of MGRN and baseline models under various adversarial and perturbation settings.
The applied data transformation methods include SwapAnt (SA), NumWord (NW), AddSent (AS), InsertAdv (IA),
AppendIrr (Al), AddPunc (AP), BackTrans (BT), TwitterType (TT), SwapNamedEnt (SN), and SwapSyn-WordNet (SW).
}
\end{table*}

\begin{table*}[t]
\centering
\renewcommand\arraystretch{0.8}
\scalebox{0.9}{
\setlength{\tabcolsep}{8.1mm}{
\begin{tabular}{lcccc}
\toprule
\midrule
\text{Case} & \text{ESIM} & \text{BERT} & \text{SyntaxBERT}& \text{MGRN}  \\
\midrule
\text{S1:}How done \textcolor{red}{you solve} this aptitude question? &
\multirow{2}{*}{\text{label:1}}& \multirow{2}{*}{\text{label:0}} & \multirow{2}{*}{\text{label:0}}& similarity:10.87\% \\
\text{S2:}How does \textcolor{blue}{I solve} aptitude questions \textcolor{blue}{on cube}? &  &   & &label:0 \\
\midrule
\text{S1:}How can I tell if \textcolor{red}{ this girl loves} me? &
\multirow{2}{*}{\text{label:1}}& \multirow{2}{*}{\text{label:1}} & \multirow{2}{*}{\text{label:1}}& similarity:12.06\% \\
\text{S2:}How can I tell if \textcolor{blue}{ this boy loves} me? &   &  &  & label:0\\
\midrule
\text{S1:}How many \textcolor{red}{12 digits number} have the sum of 4? &
\multirow{2}{*}{\text{label:1}}& \multirow{2}{*}{\text{label:1}} & \multirow{2}{*}{\text{label:1}}& similarity:18.63\% \\
\text{S2:}How many \textcolor{blue}{42 digits number} have the sum of 4? &   & 
& &label:0 \\
\midrule
\bottomrule
\end{tabular}}}
\caption{\label{citation-guide-casestudy2} The example sentence pairs of our cases. \textcolor{red}{Red} and \textcolor{blue}{Blue} are difference phrases in sentence pair. 
}
\vspace{-0.2cm}
\end{table*}

\subsection{Robustness Test Performance}

To further evaluate the robustness of our proposed MGRN framework, we conduct adversarial and perturbation-based
robustness tests on three widely studied datasets, including Quora Question Pairs (QQP), SNLI, and MNLI.
Table~\ref{citation-guide-robust} reports the performance of MGRN and several competitive baseline models
under different transformation settings. Overall, MGRN consistently achieves superior or highly competitive
performance across most perturbation types, demonstrating its strong robustness and generalization capability.
In the SwapAnt transformation, which introduces semantic contradictions by replacing words with their antonyms,
MGRN significantly outperforms all baseline models on QQP. This result highlights the advantage of explicitly
modeling fine-grained interaction patterns across multiple representation layers, enabling MGRN to better
capture semantic polarity shifts that are difficult to identify using surface-level alignments alone.
For the NumWord transformation, which requires precise numerical reasoning, all models experience performance
degradation. However, MGRN outperforms BERT by a clear margin, indicating that multi-granularity interaction
representations help preserve subtle semantic differences introduced by numerical changes.
Under synonym-based perturbations such as SwapSyn, UERBERT benefits from explicitly injected lexical knowledge
and achieves strong performance. Notably, MGRN attains comparable results without relying on external knowledge
resources, suggesting that hierarchical semantic interactions encoded in intermediate transformer layers
already provide rich contextual cues for semantic equivalence reasoning.
In transformations that disrupt surface form or syntactic regularity, such as TwitterType and AddPunc,
models that heavily depend on syntactic structures (e.g., SyntaxBERT) exhibit noticeable performance drops.
In contrast, MGRN maintains stable performance, indicating that its reasoning process relies more on semantic
interaction patterns rather than fragile surface or syntactic features.

\subsection{Case Study}

To qualitatively analyze how MGRN performs fine-grained reasoning, we present three representative examples
in Table~\ref{citation-guide-casestudy2}. These cases involve subtle lexical, gender-related, and numerical
differences that often challenge conventional NLI models.
In the first example, the non-pretrained ESIM model fails to recognize the semantic conflict caused by subtle
word substitutions, resulting in incorrect predictions. While BERT benefits from contextualized representations
and correctly predicts this case, it struggles in the third example, where numerical differences (e.g., ``12''
vs. ``42'') require precise semantic discrimination.
SyntaxBERT enhances text understanding by incorporating syntactic structures; however, in cases where sentence
pairs share highly similar syntactic patterns, such as the second and third examples, syntactic cues alone
are insufficient to distinguish semantic differences, leading to incorrect predictions.
In contrast, MGRN consistently makes correct predictions across all cases. By explicitly modeling fine-grained
interaction patterns across multiple transformer layers, MGRN is able to capture subtle semantic discrepancies
introduced by lexical substitutions, gender changes, and numerical variations. This demonstrates the advantage
of multi-granularity reasoning in accurately identifying semantic relations that cannot be resolved by
surface-level or single-layer representations alone.



\section{Conclusion}

In this paper, we proposed a novel Multi-Granularity Reasoning Network (MGRN) for natural language inference that leverages hierarchical semantic features through an interactive tensor structure. Our approach mimics the human cognitive process of language understanding, progressing from shallow to deep levels of comprehension to effectively uncover intricate semantic information behind linguistic expressions.
Extensive experiments on multiple public benchmarks demonstrate that MGRN consistently outperforms strong baseline models across various NLI tasks.

\end{document}